\documentclass[format=acmsmall, review=false, screen=true]{acmart}

\usepackage{booktabs} 
\usepackage{subfig}
\usepackage[ruled]{algorithm2e} 
\usepackage{caption}

\SetAlFnt{\small}
\SetAlCapFnt{\small}
\SetAlCapNameFnt{\small}
\SetAlCapHSkip{0pt}
\SetKwInOut{Objective}{Objective}
\IncMargin{-\parindent}

\acmJournal{TIST}
\acmVolume{10}
\acmNumber{2}
\acmArticle{12}
\acmYear{2019}
\acmMonth{2}
\copyrightyear{2019}

\setcopyright{acmlicensed}

\acmDOI{0000001.0000001}

\begin{document}
\title[Federated Machine Learning: Concept and Applications]{Federated Machine Learning: Concept and Applications}

\author{Qiang Yang}
\orcid{}
\affiliation{%
  \institution{Hong Kong University of Science and Technology}
  \streetaddress{}
  \city{}
  \state{}
  \postcode{}
  \country{Hong Kong}}
\email{}
\author{Yang Liu}
\affiliation{%
  \institution{Webank}
  \city{Shenzhen}
  \country{China}
}
\email{}
\author{Tianjian Chen}
\affiliation{%
 \institution{Webank}
 \streetaddress{}
 \city{Shenzhen}
 \state{}
 \country{China}}
\email{}
\author{Yongxin Tong}
\affiliation{%
  \institution{Beihang University}
  \streetaddress{}
  \city{Beijing}
  \state{}
  \country{China}
}
\email{}
\authorsaddresses{Authors' addresses: Qiang Yang, Hong Kong University of Science and Technology, Hong Kong, China; email: qyang@cse.ust.hk; Yang Liu, Webank, Shenzhen, China; email: yangliu@webank.com; Tianjian Chen, Webank, Shenzhen, China; email: tobychen@webank.com; Yongxin Tong (corresponding author), Advanced Innovation Center for Big Data and Brain Computing, Beihang University, Beijing, China; email: yxtong@buaa.edu.cn.}

\begin{abstract}
Today's AI still faces two major challenges. One is that in most industries, data exists in the form of isolated islands. The other is the strengthening of data privacy and security. We propose a possible solution to these challenges: secure federated learning. Beyond the federated learning framework first proposed by Google in 2016, we introduce a comprehensive secure federated learning framework, which includes horizontal federated learning, vertical federated learning and federated transfer learning. We provide definitions, architectures and applications for the federated learning framework, and provide a comprehensive survey of existing works on this subject. In addition, we propose building data networks among organizations based on federated mechanisms as an effective solution to allow knowledge to be shared without compromising user privacy.
\end{abstract}

%
%


%
%


\maketitle

\renewcommand{\shortauthors}{Q. Yang et al.}

\section{Introduction}
2016 is the year when artificial intelligence (AI) came of age. With AlphaGo\cite{44806} defeating the top human Go players, we have truly witnessed the huge potential in artificial intelligence (AI), and have began to expect more complex, cutting-edge AI technology in many applications, including driverless cars, medical care, finance, etc.  Today, AI technology is showing its strengths in almost every industry and walks of life. However, when we look back at the development of AI, it is inevitable that the development of AI has experienced several ups and downs. Will there be a next down turn for AI? When will it appear and because of what factors?  The current public interest in AI is partly driven by Big Data availability: AlphaGo in 2016 used a total of 300,000 games as training data to achieve the excellent results.

With AlphaGo's success, people naturally hope that the big data-driven AI like AlphaGo will be realized soon in all aspects of our lives. However, the real world situations are somewhat disappointing: with the exception of few industries, most fields have only limited data or poor quality data, making the realization of AI technology more difficult than we thought.  Would it be possible to fuse the data together in a common site, by transporting the data across organizations?  In fact, it is very difficult, if not impossible, in many situations to break the barriers between data sources. In general, the data required in any AI project involves multiple types. For example, in an AI-driven product recommendation service, the product seller has information about the product, data of the user's purchase, but not the data that describe  user's purchasing ability and payment habits. In most industries, data exists in the form of isolated islands. Due to industry competition, privacy security, and complicated administrative procedures, even data integration between different departments of the same company faces heavy resistance. It is almost impossible to integrate the data scattered around the country and institutions, or the cost is prohibited.

\begin{figure}
  \includegraphics[width=6cm]{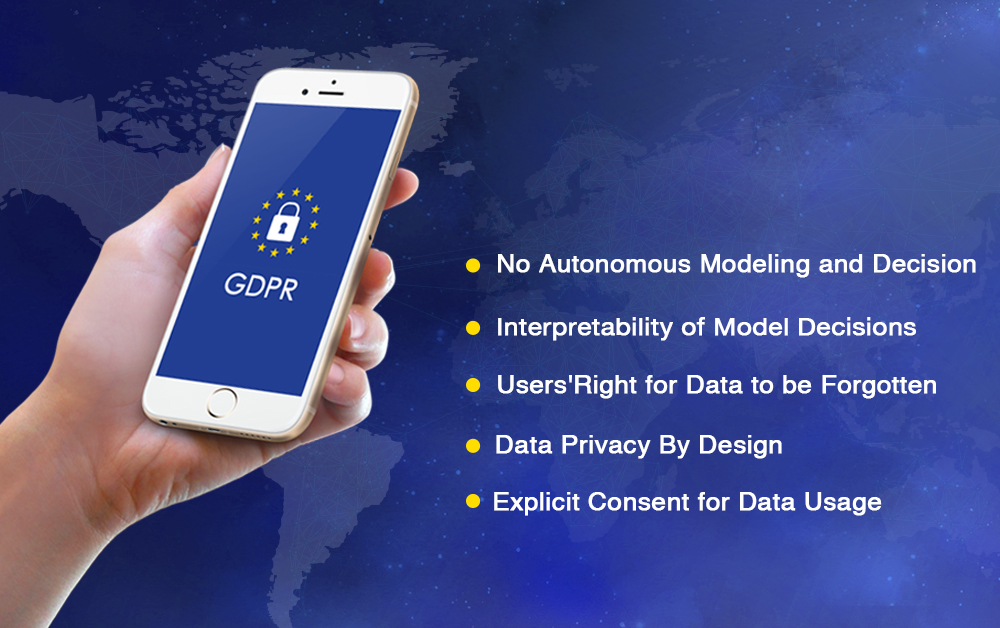}
  \caption{GDPR: EU regulation on data protection}
  \label{fig:one}
\end{figure}

At the same time, with the increasing awareness of large companies compromising on data security and user privacy, the emphasis on data privacy and security has become a worldwide major issue. News about leaks on public data are causing great concerns in public media and governments. For example, the recent data breach by Facebook has caused a wide range of protests \cite{wikifacebook}.  In response, states across the world are strengthening laws in  protection of data security and privacy.  An example is the General Data Protection Regulation (GDPR)\cite{regulation2016general}  enforced by the European Union on May 25, 2018. GDPR (Figure \ref{fig:one}) aims to protect users' personal privacy and data security. It requires businesses to use clear and plain languages for their user agreement and grants users the "right to be forgotten", that is, users can have their personal data deleted or withdrawn. Companies violating the bill will face stiff fine. Similar acts of privacy and security are being enacted in the US and China.  For example, China's Cyber Security Law and the General Principles of the Civil Law, enacted in 2017,  require that Internet businesses must not leak or tamper with the personal information that they collect and that, when conducting data transactions with third parties, they need to ensure that the proposed contract follow legal data protection obligations. The establishment of these regulations will clearly help build a more civil society, but will also pose new challenges to the  data transaction procedures commonly used today in AI.

To be more specific, traditional data processing models in AI often involves simple data transactions models, with one party collecting and transferring data to another party, and this other party will be responsible for cleaning and fusing the data.  Finally a third party will take the integrated data and build models for still other parties to use.  The models are usually the final products that are sold as a service. This traditional procedure face challenges with the above new data regulations and laws. As well,  since users may be unclear about the future uses of the models, the transactions violate laws such as the GDPR.  As a result, we face a dilemma that our data is in the form of isolated islands, but we are forbidden in many situations to collect, fuse and use the data to different places for AI processing. How to legally solve the problem of data fragmentation and isolation is a major challenge for AI researchers and practitioners today.

In this article, we give an overview of a new approach known as {\em federated learning}, which is a possible solution for these challenges. We survey existing works on federated learning, and propose definitions, categorizations and applications for a comprehensive secure federated learning framework. We discuss how the federated learning framework can be applied to various businesses successfully. In promoting federated learning, we hope to shift the focus of AI development from improving model performance, which is what most of the AI field is currently doing, to investigating methods for data integration that is compliant with data privacy and security laws.

\section{An Overview of Federated Learning}
 The concept of federated learning is proposed by Google recently \cite{DBLP:journals/corr/KonecnyMRR16,DBLP:journals/corr/McMahanMRA16,DBLP:journals/corr/KonecnyMYRSB16}. Their main idea is to build machine learning models based on data sets that are distributed across multiple devices while preventing data leakage. Recent improvements have been focusing on overcoming the statistical challenges \cite{NIPS2017_7029,zhao2018federated} and improving security \cite{Bonawitz:2017:PSA:3133956.3133982,DBLP:journals/corr/abs-1712-07557} in federated learning.  There are also research efforts to make federated learning more personalizable \cite{DBLP:journals/corr/abs-1802-07876,NIPS2017_7029}. The above works all focus on on-device federated learning where distributed mobile user interactions are involved and communication cost in massive distribution, unbalanced data distribution and device reliability are some of the major factors for optimization. In addition, data are partitioned by user Ids or device Ids, therefore, \textit{horizontally} in the data space. This line of work is very related to privacy-preserving machine learning such as  \cite{Shokri:2015:PDL:2810103.2813687} because it also considers data privacy in a decentralized collaborative learning setting. To extend the concept of federated learning to cover collaborative learning scenarios among organizations, we extend the original "federated learning" to a general concept for all privacy-preserving decentralized collaborative machine learning techniques. In \cite{CCCF-FL18}, we have given a preliminary overview of the federated learning and federated transfer learning technique. In this article, we further survey the relevant security foundations and explore the relationship with several other related areas, such as multiagent theory and privacy-preserving data mining. In this section, we provide a more comprehensive definition of federated learning which considers data partitions, security and applications. We also describe a workflow and system architecture for the federated learning system.

\subsection{Definition of Federated Learning}
Define $N$ data owners $\{\mathcal{F}_1,...\mathcal{F}_N\}$,  all of whom wish to train a machine learning model by consolidating their respective data $\{\mathcal{D}_1,...\mathcal{D}_N\}$. A conventional method is to put all data together and use $\mathcal{D} = \mathcal{D}_1\cup...\cup\mathcal{D}_N$ to train a model $\mathcal{M}_{SUM}$.  A federated learning system is a learning process in which the data owners collaboratively train a model $\mathcal{M}_{FED}$, in which process any data owner $\mathcal{F}_i$ does not expose its data $\mathcal{D}_i$ to others \footnote{Definition of data security may differ in different scenarios, but is required to provide meaning privacy guarantees. We demonstrate examples of security definitions in section 2.3}. In addition, the accuracy of $\mathcal{M}_{FED}$, denoted as $\mathcal{V}_{FED}$ should be very close to the performance of $\mathcal{M}_{SUM}$, $\mathcal{V}_{SUM}$. Formally, let $\delta$ be a non-negative real number, if
\begin{equation}\label{define}
 \mid \mathcal{V}_{FED} - \mathcal{V}_{SUM} \mid < \delta
\end{equation}
we say the federated learning algorithm has $\delta$-accuracy loss.

\subsection{Privacy of Federated Learning}
Privacy is one of the essential properties of federated learning. This requires security models and analysis to provide meaningful privacy guarantees.
In this section, we briefly review and compare different privacy techniques for federated learning, and identify approaches and potential challenges for preventing indirect leakage.

\paragraph{Secure Multi-party Computation (SMC)} SMC security models naturally involve multiple parties, and provide security proof in a well-defined simulation framework to guarantee complete zero knowledge, that is, each party knows nothing except its input and output. Zero knowledge is very desirable, but this desired property usually requires complicated computation protocols and may not be achieved efficiently. In certain scenarios, partial knowledge disclosure may be considered acceptable if security guarantees are provided. It is possible to build a security model with SMC under lower security requirement in exchange for efficiency \cite{Du2004PrivacyPreservingMS}. Recently, studies \cite{DBLP:conf/sp/MohasselZ17} used SMC framework for training machine learning models with two servers and semi-honest assumptions. Ref \cite{pmlr-v80-kilbertus18a} uses MPC protocols for model training and verification without users revealing sensitive data. One of the state-of-the-art SMC framework is Sharemind \cite{Bogdanov:2008:SFF:1462455.1462473}. Ref \cite{Mohassel:2018:AMP:3243734.3243760} proposed a 3PC model \cite{Araki:2016:HSS:2976749.2978331,Mohassel:2015:FST:2810103.2813705,cryptoeprint:2016:944} with an honest majority and consider security in both semi-honest and malicious assumptions. These works require participants' data to be secretly-shared among non-colluding servers.

\paragraph{Differential Privacy}
Another line of work use techniques Differential Privacy \cite{Dwork:2008:DPS:1791834.1791836} or k-Anonymity \cite{Sweeney:2002:KAM:774544.774552} for data privacy protection \cite{Abadi:2016:DLD:2976749.2978318,DBLP:journals/corr/abs-1710-06963,NIPS2008_3486,Song2013StochasticGD}. The methods of differential privacy, k-anonymity, and diversification \cite{Agrawal:2000:PDM:342009.335438} involve in adding noise to the data, or using generalization methods to obscure certain sensitive attributes until the third party cannot distinguish the individual, thereby making the data impossible to be restore to protect user privacy. However, the root of these methods still require that the data are transmitted elsewhere and these work usually involve a trade-off between accuracy and privacy. In \cite{DBLP:journals/corr/abs-1712-07557}, authors introduced a differential privacy approach to federated learning in order to add protection to client-side data by hiding client's contributions during training.

\paragraph{Homomorphic Encryption}
Homomorphic Encryption \cite{Rivest1978} is also adopted to protect user data privacy through parameter exchange under the encryption mechanism during machine learning \cite{CIS-274193,cryptoeprint:2017:979,Nikolaenko:2013:PRR:2497621.2498119}. Unlike differential privacy protection, the data and the model itself are not transmitted, nor can they be guessed by the other party's data. Therefore, there is little possibility of leakage at the raw data level. Recent works adopted homomorphic encryption for centralizing and training data on cloud \cite{Yuan:2014:PPB:2553575.2553597,Zhang:2016:PPD:2925261.2925353}. In practice, Additively Homomorphic Encryption \cite{Acar:2018:SHE:3236632.3214303} are widely used and polynomial approximations need to be made to evaluate non-linear functions in machine learn algorithms, resulting in the trade-offs between accuracy and privacy  \cite{Aono:2016:SSL:2857705.2857731,info:doi/10.2196/medinform.8805}.

\subsubsection{Indirect information leakage}
Pioneer works of federated learning exposes intermediate results such as parameter updates from an optimization algorithm like Stochastic Gradient Descent (SGD)  \cite{DBLP:journals/corr/McMahanMRA16,Shokri:2015:PDL:2810103.2813687}, however no security guarantee is provided and the leakage of these gradients may actually leak important data information \cite{DBLP:journals/tifs/PhongAHWM18} when exposed together with data structure such as in the case of image pixels. Researchers have considered the situation when one of the members of a federated learning system maliciously attacks others by allowing a backdoor to be inserted to learn others' data.  In \cite{bagdasaryan2018backdoor}, the authors  demonstrate that it is possible to insert hidden backdoors into a joint global model and propose a new "constrain-and-scale" model-poisoning methodology to reduce the data poisoning. In \cite{DBLP:journals/corr/abs-1805-04049}, researchers identified potential loopholes in collaborative machine learning systems, where the training data used by different parties in collaborative learning is vulnerable to inference attacks. They showed that an adversarial participant can infer membership as well as properties associated with   a subset of the training data.  They also discussed possible defenses against these attacks.In \cite{DBLP:journals/corr/abs-1804-10140}, authors expose a potential security issue associated with gradient exchanges between different parties, and  propose a secured variant of the gradient descent method and show that it tolerates up to a constant fraction of Byzantine workers.

Researchers have also started to consider blockchain as a platform for facilitating federated learning.  In \cite{kim2018ondevice}, researchers have considered a block-chained federated learning (BlockFL) architecture, where mobile devices' local learning model updates are exchanged and verified by leveraging blockchain. They have considered an optimal block generation, network scalability and robustness issues.

\begin{figure}
    \centering
	\subfloat[\scriptsize{Horizontal Federated Learning}]{

		\includegraphics[scale=0.3]{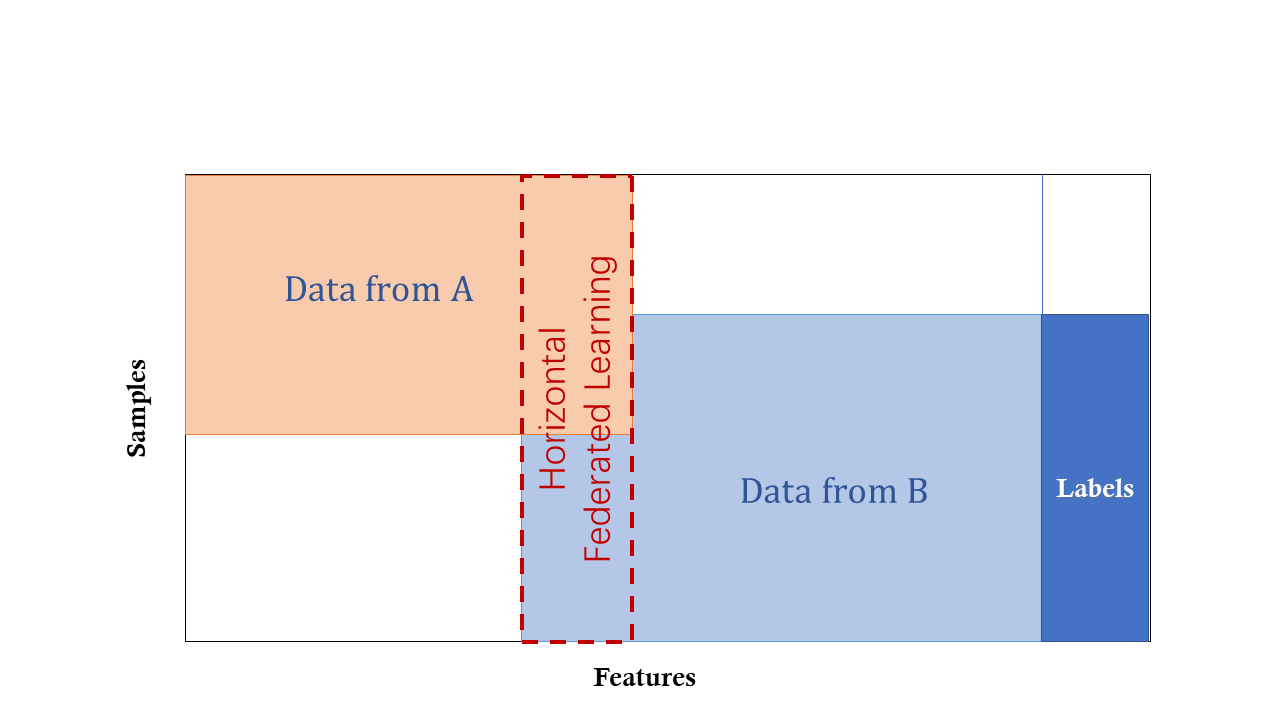}

		\label{fig:h}
    }
	
    \subfloat[\scriptsize{Vertical Federated Learning}]{
		\includegraphics[scale=0.3]{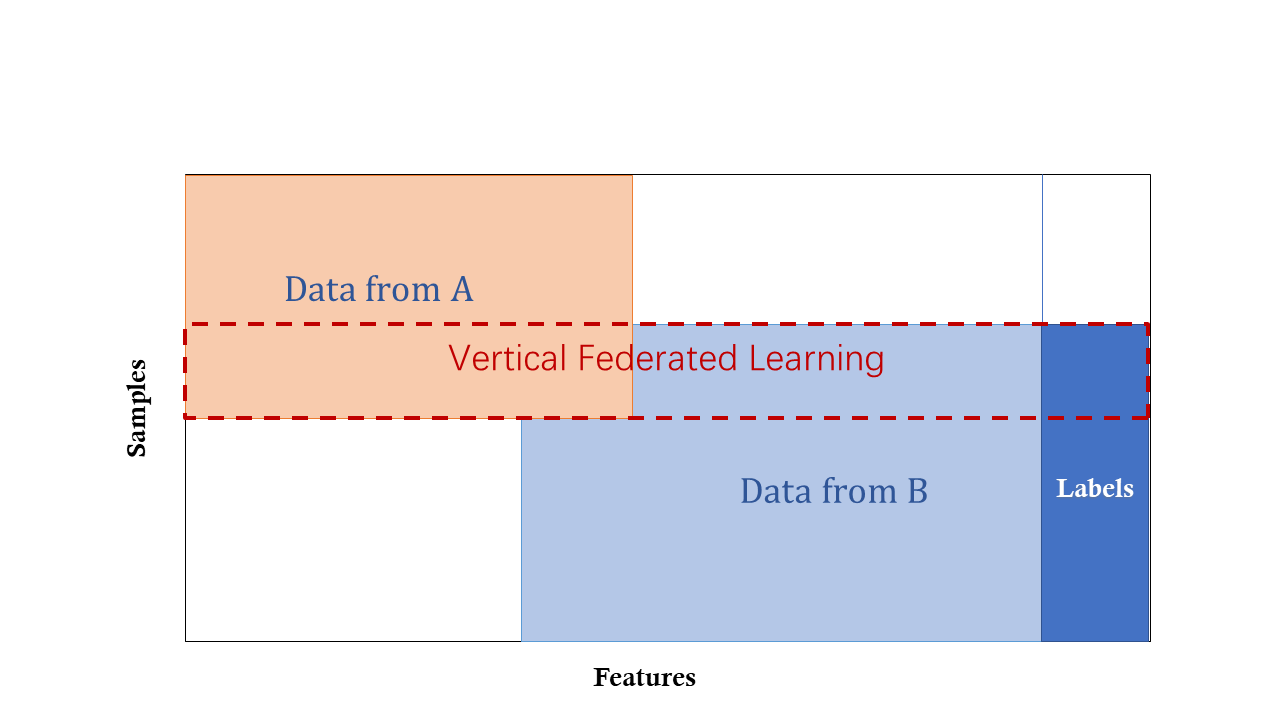}
		\label{fig:v}
		
    }

    \subfloat[\scriptsize{Federated Transfer Learning}]{

		\includegraphics[scale=0.3]{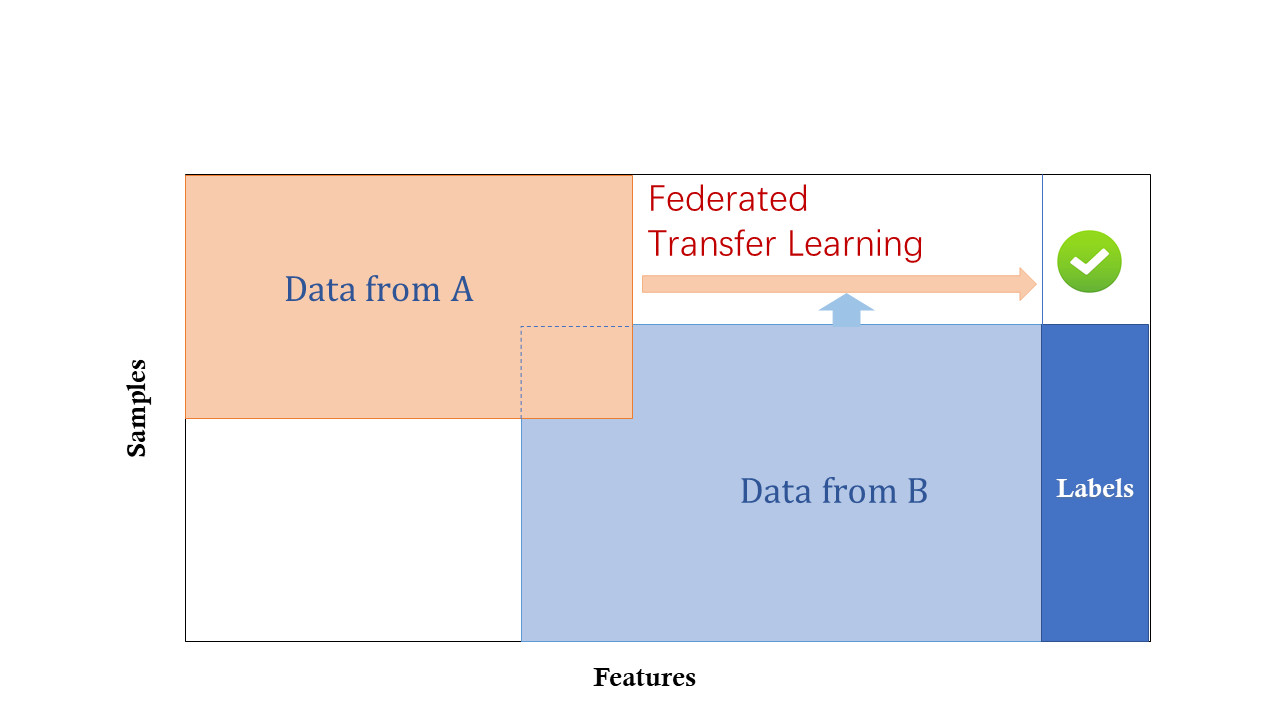}
		\label{fig:ftl}
    }
  \caption{Categorization of Federated Learning}
  \label{fig:two}
\end{figure}

\subsection{A Categorization of Federated Learning}
In this section we discuss how to categorize federated learning based on the distribution characteristics of the data.

 Let matrix $\mathcal{D}_i$ denotes the data held by each data owner $i$. Each row of the matrix represents a sample, and each column represents a feature. At the same time, some data sets may also contain label data.  We denote the features space as $\mathcal{X}$, the label space as $\mathcal{Y}$ and we use $\mathcal{I}$ to denote the sample ID space. For example, in the financial field labels may be users' credit; in the marketing field labels may be the user's purchase desire; in the education field, $\mathcal{Y}$ may be the degree of the students. The feature $\mathcal{X}$, label $\mathcal{Y}$ and sample Ids $\mathcal{I}$ constitutes the complete training dataset $(\mathcal{I},\mathcal{X}, \mathcal{Y})$. The feature and sample space of the data parties may not be identical, and we classify federated learning into horizontally federated learning, vertically federated learning and federated transfer learning based on how data is distributed among various parties in the feature and sample ID space. Figure \ref{fig:two} shows the various federated learning frameworks for a two-party scenario .

\subsubsection{Horizontal Federated Learning}
Horizontal federated learning, or sample-based federated learning, is introduced in the scenarios that data sets share the same feature space but different in samples (Figure \ref{fig:h}). For example, two regional banks may have very different user groups from their respective regions, and the intersection set of their users is very small. However, their business is very similar, so the feature spaces are the same. Ref \cite{Shokri:2015:PDL:2810103.2813687} proposed a collaboratively deep learning scheme where participants train independently and share only subsets of updates of parameters. In 2017, Google proposed a horizontal federated learning solution for Android phone model updates \cite{DBLP:journals/corr/McMahanMRA16}. In that framework, a single user using an Android phone updates the model parameters locally and uploads the parameters to the Android cloud, thus jointly training the centralized model together with other data owners. A secure aggregation scheme to protect the privacy of aggregated user updates under their federated learning framework is also introduced \cite{Bonawitz:2017:PSA:3133956.3133982}. Ref \cite{DBLP:journals/tifs/PhongAHWM18} uses additively homomorphic encryption for model paramter aggregation to provide security against the central server.

In \cite{NIPS2017_7029}, a multi-task style federated learning system is proposed to allow multiple sites to complete separate tasks, while sharing knowledge and preserving security.  Their proposed multi-task learning model can in addition address high communication costs, stragglers, and fault tolerance issues.  In \cite{DBLP:journals/corr/McMahanMRA16}, the authors proposed to build a secure client-server structure where the federated learning system partitions data by users, and allow models built at client devices to collaborate at the server site to build a global federated model.  The process of model building ensures that there is no data leakage.  Likewise, in \cite{DBLP:journals/corr/KonecnyMRR16}, the authors proposed  methods to improve the communication cost to facilitate the training of centralized models based on data distributed over mobile clients. Recently, a compression approach called Deep Gradient Compression \cite{DBLP:journals/corr/abs-1712-01887} is proposed to greatly reduce the communication bandwidth in large-scale distributed training.

We summarize horizontal federated learning as:
\begin{equation}
    \mathcal{X}_i = \mathcal{X}_j,\ \ \mathcal{Y}_i = \mathcal{Y}_j, \ \  I_i\neq I_j,\ \ \forall \mathcal{D}_i, \mathcal{D}_j, i\neq j
\end{equation}

\paragraph{Security Definition}  A horizontal federated learning system typically assumes honest participants and security against a honest-but-curious server \cite{DBLP:journals/tifs/PhongAHWM18,Bonawitz:2017:PSA:3133956.3133982}. That is, only the server can compromise the privacy of data participants. Security proof has been provided in these works. Recently another security model considering malicious user \cite{DBLP:journals/corr/HitajAP17} is also proposed, posing additional privacy challenges. At the end of the training, the universal model and the entire model parameters are exposed to all participants.

\subsubsection{Vertical Federated Learning}
Privacy-preserving machine learning algorithms have been proposed for vertically partitioned data, including Cooperative Statistical Analysis \cite{Du:2001:PCS:872016.872181}, association rule mining \cite{Vaidya:2002:PPA:775047.775142}, secure linear regression \cite{Karr2004PrivacyPreservingAO,Sanil:2004:PPR:1014052.1014139,Gascn2016SecureLR}, classification \cite{Du2004PrivacyPreservingMS} and gradient descent \cite{Wan:2007:PGD:1281192.1281275}. Recently, Ref \cite{Hardy2017PrivateFL,DBLP:journals/corr/abs-1803-04035} proposed a vertical federated learning scheme to train a privacy-preserving logistic regression model. The authors studied the effect of entity resolution on the learning performance and applied Taylor approximation to the loss and gradient functions so that homomorphic encryption can be adopted for privacy-preserving computations.

Vertical federated learning  or feature-based federated learning (Figure \ref{fig:v}) is applicable to the cases that two data sets share the same sample ID space but differ in feature space. For example, consider two different companies in the same city, one is a bank, and the other is an e-commerce company. Their user sets are likely to contain most of the residents of the area, so the intersection of their user space is large. However, since the bank records the user's revenue and expenditure behavior and credit rating, and the e-commerce retains the user's browsing and purchasing history, their feature spaces are very different. Suppose that we want both parties to have a prediction model for product purchase based on user and product information.

Vertically federated learning is the process of aggregating these different features and computing the training loss and gradients in a privacy-preserving manner to build a model with data from both parties collaboratively. Under such a federal mechanism, the identity and the status of each participating party is the same, and the federal system helps everyone establish a "common wealth" strategy, which is why this system is called "federated learning.". Therefore, in such a system, we have:
\begin{equation}
    \mathcal{X}_i \neq \mathcal{X}_j,\ \ \mathcal{Y}_i \neq \mathcal{Y}_j, \ \  I_i = I_j\ \ \forall \mathcal{D}_i, \mathcal{D}_j, i\neq j
\end{equation}

\paragraph{Security Definition}  A vertical federated learning system typically assumes honest-but-curious participants. In a two-party case, for example, the two parties are non-colluding and at most one of them are compromised by an adversary. The security definition is that the adversary can only learn data from the client that it corrupted but not data from the other client beyond what is revealed by the input and output. To facilitate the secure computations between the two parties, sometimes a Semi-honest Third Party (STP) is introduced, in which case it is assumed that STP does not collude with either party. SMC provides formal privacy proof for these protocols \cite{Goldreich:1987:PAM:28395.28420}. At the end of learning, each party only holds the model parameters associated to its own features, therefore at inference time, the two parties also need to collaborate to generate output.

\subsubsection{Federated Transfer Learning (FTL)}
Federated Transfer Learning applies to the scenarios that the two data sets differ not only in samples but also in feature space. Consider two institutions, one is a bank located in China, and the other is an e-commerce company located in the United States. Due to geographical restrictions, the user groups of the two institutions have a small intersection. On the other hand, due to the different businesses, only a small portion of the feature space from both parties overlaps. In this case, transfer learning \cite{Pan:2010:STL:1850483.1850545} techniques can be applied to provide solutions for the entire sample and feature space under a federation (Figure\ref{fig:ftl}). Specially, a common representation between the two feature space is learned using the limited common sample sets and later applied to obtain predictions for samples with only one-side features. FTL is an important extension to the existing federated learning systems because it deals with problems exceeding the scope of existing federated learning algorithms:
\begin{equation}
    \mathcal{X}_i \neq \mathcal{X}_j,\ \ \mathcal{Y}_i \neq \mathcal{Y}_j, \ \  I_i \neq I_j\ \ \forall \mathcal{D}_i, \mathcal{D}_j, i\neq j
\end{equation}

\paragraph{Security Definition}  A  federated transfer learning system typically involves two parties. As will be shown in the next section, its protocols are similar to the ones in vertical federated learning, in which case  the security definition for vertical federated learning can be extended here.

\begin{figure}
  \includegraphics[width=13cm]{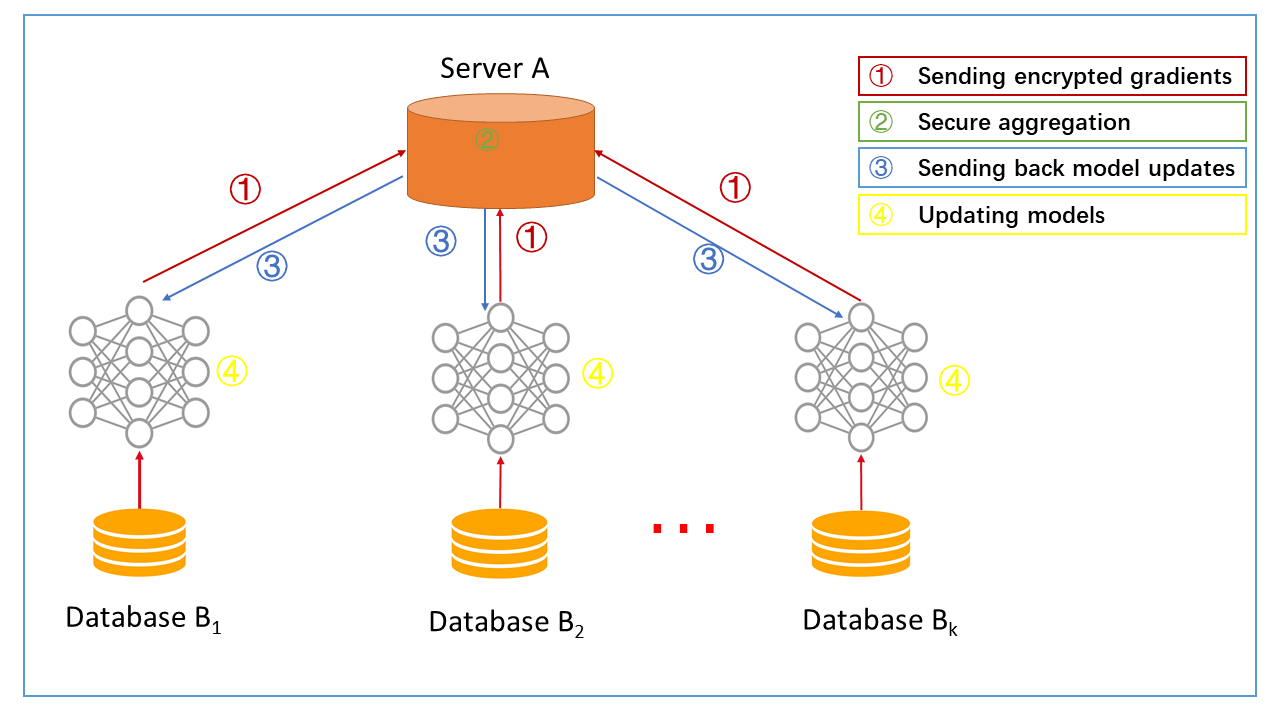}
  \caption{Architecture for a horizontal federated learning system}
  \label{fig:four}
\end{figure}

\subsection{Architecture for a federated learning system}

In this section, we illustrate examples of general architectures for a federated learning system. Note that the architectures of horizontal and vertical federated learning systems are quite different by design, and we will introduce them separately.

\subsubsection{Horizontal Federated Learning}
A typical architecture for a horizontal federated learning system is shown in Figure \ref{fig:four}.  In this system, k participants with the same data structure collaboratively learn a machine learning model with the help of a parameter or cloud server. A typical assumption is that the participants are honest whereas the server is honest-but-curious, therefore no leakage of information from any participants to the server is allowed \cite{DBLP:journals/tifs/PhongAHWM18}. The training process of such a system usually contain the following four steps:
\begin{itemize}
	\item \textbf{Step 1}: participants locally compute training gradients, mask a selection of gradients with encryption \cite{DBLP:journals/tifs/PhongAHWM18}, differential privacy \cite{Shokri:2015:PDL:2810103.2813687} or secret sharing \cite{Bonawitz:2017:PSA:3133956.3133982} techniques, and send masked results to server;
	\item \textbf{Step 2}: Server performs secure aggregation without learning information about any participant;
	\item \textbf{Step 3}: Server send back the aggregated results to participants;
	\item \textbf{Step 4}: Participants update their respective model with the decrypted gradients.
\end{itemize}
Iterations through the above steps continue until the loss function converges, thus completing the entire training process. This architecture is independent of specific machine learning algorithms (logistic regression, DNN etc) and all participants will share the final model parameters.

\paragraph{Security Analysis} The above architecture is proved to protect data leakage against the semi-honest server, if gradients aggregation is done with SMC \cite{Bonawitz:2017:PSA:3133956.3133982} or Homomorphic Encryption \cite{DBLP:journals/tifs/PhongAHWM18}. But it may be subject to attack in another security model by a malicious participant training a Generative Adversarial Network (GAN) in the collaborative learning process \cite{DBLP:journals/corr/HitajAP17}.

\begin{figure}
  \includegraphics[width=12.5cm]{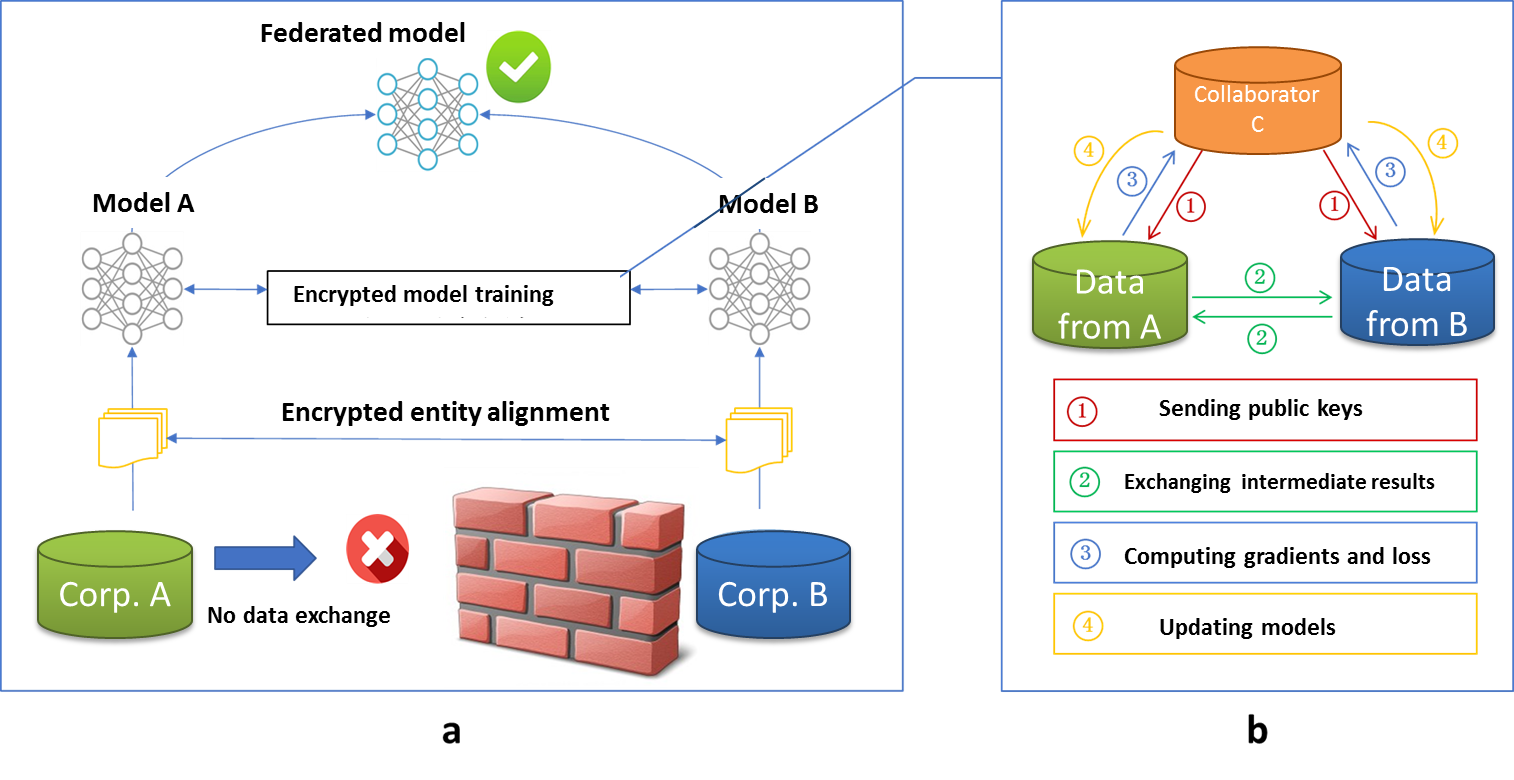}
  \caption{Architecture for a vertical federated learning system}
  \label{fig:three}
\end{figure}

\subsubsection{Vertical Federated Learning}
Suppose that companies A and B would like to jointly train a machine learning model, and their business systems each have their own data. In addition, Company B also has label data that the model needs to predict. For data privacy and security reasons, A and B cannot directly exchange data. In order to ensure the confidentiality of the data during the training process, a third-party collaborator C is involved. Here we assume the collaborator C is honest and does not collude with A or B, but party A and B are honest-but-curious to each other. A trusted third party C a reasonable assumption since party C can be played by authorities such as governments or replaced by secure computing node such as Intel Software Guard Extensions (SGX) \cite{DBLP:conf/fc/BahmaniBBPSSW17}. The federated learning system consists of two parts, as shown in Figure \ref{fig:three}.
\paragraph{Part 1} Encrypted entity alignment. Since the user groups of the two companies are not the same, the system uses the encryption-based user ID alignment techniques such as \cite{liang2004privacy,Scannapieco:2007:PPS:1247480.1247553} to confirm the common users of both parties without A and B exposing their respective data. During the entity alignment, the system does not expose users that do not overlap with each other.
\paragraph{Part 2} Encrypted model training. After determining the common entities, we can use these common entities' data to train the machine learning model. The training process can be divided into the following four steps  (as shown in Figure \ref{fig:three}):
\begin{itemize}
	\item \textbf{Step 1}: collaborator C creates encryption pairs, send public key to A and B;
	\item \textbf{Step 2}: A and B encrypt and exchange the intermediate results for gradient and loss calculations;
	\item \textbf{Step 3}: A and B computes encrypted gradients and adds \textit{additional mask}, respectively,and B also computes encrypted loss; A and B send encrypted values to C;
	\item \textbf{Step 4}: C decrypts and send the decrypted gradients and loss back to A and B;  A and B unmask the gradients, update the model parameters accordingly.
\end{itemize}
Here we illustrate the training process using linear regression and homomorphic encryption as an example. To train a linear regression model with gradient descent methods, we need secure computations of its loss and gradients. Assuming learning rate $\eta$, regularization parameter $\lambda$, data set $\{x_i^A\}_{i \in \mathcal{D}_A}$, $\{x_i^B, y_i\}_{i \in \mathcal{D}_B}$, and model paramters  $\Theta_A$, $\Theta_B$ corresponding to the feature space of $x_i^A$, $x_i^B$ respectively, the training objective is:
\begin{equation}\label{obj}
    \underset{\Theta_A,\Theta_B}{\operatorname{min}}\sum_i{||\Theta_Ax_i^A+\Theta_Bx_i^B-y_i||^2 + \frac{\lambda}{2}(||\Theta_A||^2 +||\Theta_B||^2)}
\end{equation}
let $u_i^A = \Theta_A x_i^A$, $u_i^B =\Theta_B x_i^B$, the encrypted loss is:
\begin{equation}\label{loss}
       [[\mathcal{L}]] = [[\sum_i((u_i^A+u_i^B-y_i))^2+  \frac{\lambda}{2}(||\Theta_A||^2+||\Theta_B||^2)]]\\
\end{equation}
where additive homomorphic encryption is denoted as $[[\cdot]]$. Let $[[\mathcal{L}_A]] = [[\sum_i((u_i^A)^2)+\frac{\lambda}{2}\Theta_A^2]]$, $[[\mathcal{L}_B]]= [[\sum_i((u_i^B-y_i)^2)+\frac{\lambda}{2}\Theta_B^2]]$, and $[[\mathcal{L}_{AB}]] = 2\sum_i([[u_i^A]](u_i^B-y_i))$, then
\begin{equation}\label{loss2}
       [[\mathcal{L}]] = [[\mathcal{L}_A]] + [[\mathcal{L}_B]] + [[\mathcal{L}_{AB}]]
\end{equation}
Similarly, let $[[d_i]] = [[u_i^A]] +[[u_i^B-y_i]]$, then gradients are:

\begin{equation}\label{ga}
    [[\frac{\partial\mathcal{L}}{\partial\Theta_A}]]= \sum_i{[[d_i]]x_i^A} + [[\lambda\Theta_A]]
\end{equation}

\begin{equation}\label{gb}
    [[\frac{\partial\mathcal{L}}{\partial\Theta_B}]]= \sum_i{[[d_i]]x_i^B} + [[\lambda\Theta_B]]
\end{equation}

\begin{table}[ht]
 \caption{Training Steps for Vertical Federated Learning : Linear Regression}
    \centering
    \begin{tabular}{p{2cm}|p{3.5cm}|p{3.5cm}|p{3.5cm}}
    \hline
          &  party A &  party B & party C\\
          \hline
          \hline
         step 1 & initialize $\Theta_A$ & initialize $\Theta_B$ & create an encryption key pair, send public key to A and B; \\
         & & & \\
         step 2 &  compute $[[u_i^A]]$,$[[\mathcal{L}_{A}]]$ and send to B; & compute $[[u_i^B]]$,$[[d_i^B]]$,$[[\mathcal{L}]]$, send $[[d_i^B]]$ to A,  send $[[\mathcal{L}]]$ to C; & \\
         & & & \\
         step 3 & initialize $R_A$, compute $[[\frac{\partial\mathcal{L}}{\partial\Theta_A}]] +[[R_A]]$ and send to C; & initialize $R_B$, compute  $[[\frac{\partial\mathcal{L}}{\partial\Theta_B}]] +[[R_B]]$ and send to C; & C decrypt $\mathcal{L}$,  send $\frac{\partial\mathcal{L}}{\partial\Theta_A} +R_A$ to A, $\frac{\partial\mathcal{L}}{\partial\Theta_B} +R_B$ to B;\\
         & & & \\
         step 4 & update $\Theta_A$ & update $\Theta_B$ & \\
         \hline
         \hline
         what is obtained & $\Theta_A$ & $\Theta_B$ & \\
         \hline
    \end{tabular}
    \label{tab:LR}
\end{table}

\begin{table}[ht]
 \caption{Evaluation Steps for Vertical Federated Learning : Linear Regression}
    \centering
    \begin{tabular}{p{2cm}|p{3.5cm}|p{3.5cm}|p{3.5cm}}
    \hline
          &  party A &  party B & inquisitor C \\
          \hline
          \hline
         step 0 &  &  & send user ID $i$ to A and B; \\
            & & & \\
         step 1 & compute $u_i^A$ and send to C & compute $u_i^B$ and send to C; & get result $u_i^A+u_i^B$; \\
      \hline
    \end{tabular}
    \label{tab:LReval}
\end{table}

See Table \ref{tab:LR} and \ref{tab:LReval} for the detailed steps. During entity alignment and model training, the data of A and B are kept locally, and the data interaction in training does not lead to data privacy leakage. Note potential information leakage to C may or may not be considered to be privacy violation. To further prevent C to learn information from A or B in this case, A and B can further hide their gradients from C by adding encrypted random masks. Therefore, the two parties achieve training a common model cooperatively with the help of federated learning. Because during the training, the loss and gradients each party receives are exactly the same as the loss and gradients they would receive if jointly building a model with data gathered at one place without privacy constraints, that is, this model is lossless. The efficiency of the model depends on the communication cost and computation cost of encrypted data. In each iteration the information sent between A and B scales with the number of overlapping samples. Therefore the efficiency of this algorithm can be further improved by adopting distributed parallel computing techniques.

\paragraph{Security Analysis} The training protocol shown in Table \ref{tab:LR} does not reveal any information to C, because all C learns are the masked gradients and the randomness and secrecy of the masked matrix are guaranteed \cite{Du2004PrivacyPreservingMS}. In the above protocol, party A learns its gradient at each step, but this is not enough for A to learn any information from B according to equation \ref{ga}, because the security of scalar product protocol is well-established based on the inability of solving n equations in more than n unknowns \cite{Du2004PrivacyPreservingMS,Vaidya:2002:PPA:775047.775142}. Here we assume the number of samples $N_A$ is much greater than $n_A$, where $n_A$ is the number of features. Similarly, party B can not learn any information from A. Therefore the security of the protocol is proved. Note we have assumed that both parties are semi-honest. If a party is malicious and cheats the system by faking its input, for example, party A submits only one non-zero input with only one non-zero feature, it can tell the value of $u_i^B$ for that feature of that sample. It still can not tell $x_i^B$ or $\Theta_B$ though, and the deviation will distort results for the next iteration, alarming the other party who will terminate the learning process. At the end of the training process, each party (A or B) remains oblivious to the data structure of the other party, and it obtains the model parameters associated only with its own features. At inference time, the two parties need to collaboratively compute the prediction results, with the steps shown in Table \ref{tab:LReval}, which still do not lead to information leakage.

\subsubsection{Federated Transfer Learning}
Suppose in the above vertical federated learning example, party A and B only have a very small set of overlapping samples and we are interested in learning the labels for all the data set in party A. The architecture described in the above section so far only works for the overlapping data set. To extend its coverage to the entire sample space, we introduce transfer learning. This does not change the overall architecture shown in Figure \ref{fig:three} but the details of the intermediate results that are exchanged between party A and party B. Specifically, transfer learning typically involves in learning a common representation between the features of party A and B, and minimizing the errors in predicting the labels for the target-domain party by leveraging the labels in the source-domain party (B in this case). Therefore the gradient computations for party A and party B are different from that in the vertical federated learning scenario. At inference time, it still requires both parties to compute the prediction results.

\subsubsection{Incentives Mechanism}
In order to fully commercialize federated learning among different organizations, a fair platform and incentive mechanisms needs to be developed \cite{Faltings:2017:GTD:3158272}. After the model is built, the performance of the model will be manifested in the actual applications and this performance can be recorded in a permanent data recording mechanism (such as Blockchain). Organizations that provide more data will be better off, and the model's effectiveness depends on the data provider's contribution to the system. The effectiveness of these models are distributed to parties based on federated mechanisms and continue to motivate more organizations to join the data federation.

The implementation of the above architecture not only considers the privacy protection and effectiveness of collaboratively-modeling among multiple organizations, but also considers how to reward organizations that contribute more data, and how to implement incentives with a consensus mechanism. Therefore, federated learning is a "closed-loop" learning mechanism.

\section{Related Works}
\label{}
Federated learning enables multiple parties to collaboratively construct a machine learning model while keeping their private training data private. As a novel technology, federated learning has several threads of originality, some of which are rooted on existing fields. Below we explain the relationship between federated learning and other related concepts from multiple perspectives.

\subsection{Privacy-preserving machine learning}

Federated learning can be considered as privacy-preserving decentralized collaborative machine learning, therefore it is tightly related to multi-party privacy-preserving machine learning. Many research efforts have been devoted to this area in the past. For example, Ref \cite{Du:2002:BDT:850782.850784, 10.1007/11535706_11} proposed algorithms for secure multi-party decision tree for vertically partitioned data. Vaidya and Clifton proposed secure association mining rules \cite{Vaidya:2002:PPA:775047.775142}, secure k-means \cite{Vaidya:2003:PKM:956750.956776}, Naive Bayes classifier \cite{Vaidya_privacypreserving} for vertically partitioned data. Ref \cite{Kantarcioglu:2004:PDM:1018031.1018322} proposed an algorithm for association rules on horizontally partitioned data.
Secure Support Vector Machines algorithms are developed for vertically partitioned data  \cite{Yu:2006:PSU:1141277.1141415} and horizontally partitioned data \cite{Yu:2006:PSC:2097044.2097132}. Ref \cite{Du2004PrivacyPreservingMS} proposed secure protocols for multi-party linear regression and classification.  Ref \cite{Wan:2007:PGD:1281192.1281275} proposed secure multi-party gradient descent methods.  The above works all used secure multi-party computation (SMC) \cite{Yao:1982:PSC:1382436.1382751,Goldreich:1987:PAM:28395.28420} for privacy guarantees.

Nikolaenko et al.\cite{Nikolaenko:2013:PRR:2497621.2498119} implemented a privacy-preserving protocol for linear regression on horizontally partitioned data using homomorphic encryption and Yao's garbled circuits and Ref  \cite{Gascn2016SecureLR, cryptoeprint:2017:979} proposed a linear regression approach for vertically partitioned data. These systems solved the linear regression problem directly. Ref \cite{DBLP:journals/iacr/MohasselZ17} approached the problem with Stochastic Gradient Descent (SGD) and they also proposed privacy-preserving protocols for logistic regression and neural networks. Recently, a follow-up work with a three-server model is proposed \cite{Mohassel:2018:AMP:3243734.3243760}. Aono et al.\cite{Aono:2016:SSL:2857705.2857731} proposed a secure logistic regression protocol using homomorphic encryption. Shokri and Shmatikov \cite{Shokri:2015:PDL:2810103.2813687} proposed training of neural networks for horizontally partitioned data with exchanges of updated parameters. Ref \cite{DBLP:journals/tifs/PhongAHWM18} used the additively homomorphic encryption to preserve the privacy of gradients and enhance the security of the system.
With the recent advances in deep learning, privacy-preserving neural networks inference is also receiving a lot of research interests\cite{DBLP:journals/corr/abs-1711-05189,cryptonets-applying-neural-networks-to-encrypted-data-with-high-throughput-and-accuracy,Chabanne2017PrivacyPreservingCO,Bourse2017FastHE,DBLP:journals/corr/RouhaniRK17,Liu:2017:ONN:3133956.3134056,DBLP:journals/corr/abs-1801-03239}.

\subsection{Federated Learning vs Distributed Machine Learning}

Horizontal federated learning at first sight is somewhat similar to Distributed Machine Learning. Distributed machine learning covers many aspects, including distributed storage of training data, distributed operation of computing tasks, distributed distribution of model results, etc. Parameter Server \cite{Ho:2013:MED:2999611.2999748} is a typical element in distributed machine learning. As a tool to accelerate the training process, the parameter server stores data on distributed working nodes, allocates data and computing resources through a central scheduling node, so as to train the model more efficiently. For horizontally federated learning, the working node represents the data owner. It has full autonomy for the local data, and can decide when and how to join the federated learning. In the parameter server, the central node always takes the control, so federated learning is faced with a more complex learning environment. Secondly, federated learning emphasizes the data privacy protection of the data owner during the model training process. Effective measures to protect data privacy can better cope with the increasingly stringent data privacy and data security regulatory environment in the future.

Like in distributed machine learning settings, federated learning will also need to address Non-IID data.  In \cite{zhao2018federated} showed that with non-iid local data, performance can be greatly reduced for federated learning.  The authors in response supplied a new method to address the issue similar to transfer learning.

\subsection{Federated Learning vs Edge Computing}
Federated learning can be seen as an operating system for edge computing, as it provides the learning protocol for coordination and security.  In \cite{DBLP:journals/corr/abs-1804-05271}, authors considered generic class of machine learning models that are trained using gradient-descent based approaches. They analyze the convergence bound of distributed gradient descent from a theoretical point of view, based on which they propose a control algorithm that determines the best trade-off between local update and global parameter aggregation to minimize the loss function under a given resource budget.

\subsection{Federated Learning vs Federated Database Systems}
Federated Database Systems \cite{Sheth:1990:FDS:96602.96604} are systems that integrate multiple database units and manage the integrated system as a whole. The federated database concept is proposed to achieve interoperability with multiple independent databases. A federated database system often uses distributed storage for database units, and in practice the data in each database unit is heterogeneous. Therefore, it has many similarities with federated learning in terms of the type and storage of data. However, the federated database system does not involve any privacy protection mechanism in the process of interacting with each other, and all database units are completely visible to the management system. In addition, the focus of the federated database system is on the basic operations of data including inserting, deleting, searching, and merging, etc., while the purpose of federated learning is to establish a joint model for each data owner under the premise of protecting data privacy, so that the various values and laws the data contain serve us better.

\section{Applications}

As an innovative modeling mechanism that could train a united model on data from multiple parties without compromising privacy and security of those data, federated learning has a promising application in sales, financial, and many other industries, in which data cannot be directly aggregated for training machine learning models due to factors such as intellectual property rights, privacy protection, and data security.

Take the smart retail as an example. Its purpose is to use machine learning techniques to provide customers with personalized services, mainly including product recommendation and sales services. The data features involved in the smart retail business mainly include user purchasing power, user personal preference, and product characteristics. In practical applications, these three data features are likely to be scattered among three different departments or enterprises. For example, a user's purchasing power can be inferred from her bank savings and her personal preference can be analyzed from her social networks, while the characteristics of products are recorded by an e-shop. In this scenario, we are facing two problems. First, for the protection of data privacy and data security, data barriers between banks, social networking sites, and e-shopping sites are difficult to break. As a result, data cannot be directly aggregated to train a model. Second, the data stored in the three parties are usually heterogeneous, and traditional machine learning models cannot directly work on heterogeneous data. For now, these problems have not been effectively solved with traditional machine learning methods, which hinder the popularization and application of artificial intelligence in more fields.

Federated learning and transfer learning are the key to solving these problems.  First, by exploiting the characteristics of federated learning, we can build a machine learning model for the three parties without exporting the enterprise data, which not only fully protects data privacy and data security, but also provides customers with personalized and targeted services and thereby achieves mutual benefits. Meanwhile, we can leverage transfer learning to address the data heterogeneity problem and break through the limitations of traditional artificial intelligence techniques. Therefore federated learning provides a good technical support for us to build a cross-enterprise, cross-data, and cross-domain ecosphere for big data and artificial intelligence.

One can use federated learning framework for multi-party database querying without exposing the data.  For example, supposed in a finance application we are interested in detecting multi-party borrowing, which has been a major risk factor in the banking industry.  This happens when certain users maliciously borrows from one bank to pay for the loan at another bank. Multi-party borrowing is a threat to financial stability as a large number of such illegal actions may cause the entire financial system to collapse. To find such users without exposing the user list to each other between banks $A$ and $B$, we can exploit a federated learning framework.  In particular, we can use the encryption mechanism of federated learning and encrypt the user list at each party, and then take the intersection of the encrypted list in the federation.  The decryption of the final result gives the list of multi-party borrowers, without exposing the other "good" users to the other party.  As we will see below, this operation corresponds to the vertical federated learning framework.

Smart healthcare is another domain which we expect will greatly benefit from the rising of federated learning techniques. Medical data such as disease symptoms, gene sequences, medical reports are very sensitive and private, yet medical data are difficult to collect and they exist in isolated medical centers and hospitals. The insufficiency of data sources and the lack of labels have led to an unsatisfactory performance of machine learning models, which becomes the bottleneck of current smart healthcare. We envisage that if all medical institutions are united and share their data to form a large medical dataset, then the performance of machine learning models trained on that large medical dataset would be significantly improved. Federated learning combining with transfer learning is the main way to achieve this vision. Transfer learning could be applied to fill the missing labels thereby expanding the scale of the available data and further improving the performance of a trained model. Therefore, federated transfer learning would play a pivotal role in the development of smart healthcare and it may be able to take human health care to a whole new level.

\begin{figure}
  \includegraphics[width=6cm]{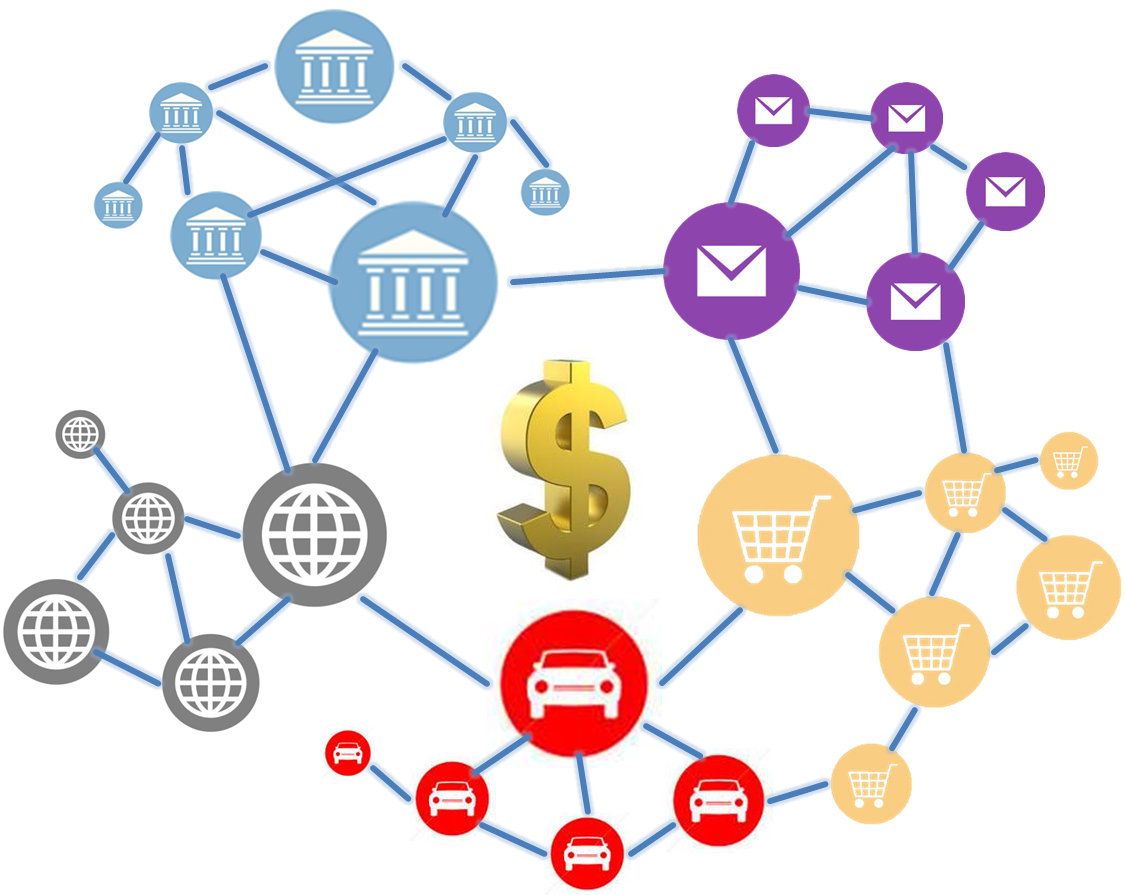}
  \caption{Data Alliance Allocates the Benefits on Blockchain}
  \label{fig:fr}
\end{figure}

\section{Federated Learning and Data Alliance of Enterprises}
Federated learning is not only a technology standard but also a business model. When people realize the effects of big data, the first thought that occurs to them is to aggregate the data together, compute the models through a remote processor and then download the results for further use. Cloud computing comes into being under such demands. However, with the increasing importance of data privacy and data security and a closer relationship between a company's profits and its data, the cloud computing model has been challenged.
However, the business model of federated learning has provided a new paradigm for applications of big data. When the isolated data occupied by each institution fails to produce an ideal model, the mechanism of federated learning makes it possible for institutions and enterprises to share a united model without data exchange. Furthermore, federated learning could make equitable rules for profits allocation with the help of consensus mechanism from blockchain techniques. The data possessors, regardless of the scale of data they have, will be motivated to join in the data alliance and make their own profits. We believe that the establishment of the business model for data alliance and the technical mechanism for federated learning should be carried out together. We would also make standards for federated learning in various fields to put it into use as soon as possible.

\section{Conclusions and Prospects}
In recent years, the isolation of data and the emphasis on data privacy are becoming the next challenges for artificial intelligence, but federated learning has brought us new hope. It could establish a united model for multiple enterprises while the local data is protected, so that enterprises could win together taking the data security as premise. This article generally introduces the basic concept, architecture and techniques of federated learning, and discusses its potential in various applications. It is expected that in the near future, federated learning would break the barriers between industries and establish a community where data and knowledge could be shared together with safety, and the benefits would be fairly distributed according to the contribution of each participant. The bonus of artificial intelligence would finally be brought to every corner of our lives.

\small
\bibliographystyle{ACM-Reference-Format}


\end{document}